\documentclass{article}

\usepackage[final]{neurips_2023_ml4ps}

\usepackage[utf8]{inputenc} 
\usepackage[T1]{fontenc}    
\usepackage{hyperref}       
\usepackage{url}            
\usepackage{booktabs}       
\usepackage{amsfonts}       
\usepackage{nicefrac}       
\usepackage{microtype}      
\usepackage[svgnames]{xcolor}         
\usepackage{graphicx}
\usepackage{soul}  

\title{PACuna: Automated Fine-Tuning of Language Models for Particle Accelerators}

\author{%
  Antonin Sulc\thanks{\href{https://github.com/sulcantonin/LLM_NeuralIPS23}{https://github.com/sulcantonin/LLM\_NeuralIPS23.git}} \\
  DESY,\\
  Hamburg, Germany \\
  \texttt{antonin.sulc@desy.de} \\
  \And
  Raimund Kammering\\
  DESY,\\
  Hamburg, Germany \\
  \And
  Annika Eichler\\
  DESY,\\
  Hamburg, Germany \\
  \And
  Tim Wilksen\\
  DESY,\\
  Hamburg, Germany
}

\newcommand{\sect}[1]{Section~\ref{#1}}
\newcommand{\fig}[1]{Figure~\ref{#1}}
\newcommand{\chat}[1]{\textit{#1}}
\newcommand{\chname}[1]{\texttt{#1}}

\newcommand{\T}{$\mathcal{T}$}
\newcommand{\Tqa}{$\mathcal{T}_{qa}$}
\newcommand{\Tp}{$\mathcal{T}_{p}$}
\newcommand{\correct}[1]{{\color{blue}\underline{#1}}}
\newcommand{\wrong}[1]{{\color{red}\underline{#1}}}

\newcommand{\answerYes}[1][]{\textcolor{blue}{[Yes] #1}}
\newcommand{\answerNo}[1][]{\textcolor{orange}{[No] #1}}
\newcommand{\answerNA}[1][]{\textcolor{gray}{[N/A] #1}}

\begin{document}

\maketitle

\begin{abstract}
Navigating the landscape of particle accelerators has become increasingly challenging with recent surges in contributions. These intricate devices challenge comprehension, even within individual facilities. To address this, we introduce PACuna, a fine-tuned language model refined through publicly available accelerator resources like conferences, pre-prints, and books. We automated data collection and question generation to minimize expert involvement and make the code available. PACuna demonstrates proficiency in addressing accelerator questions validated by experts. Our approach shows adapting language models to scientific domains by fine-tuning technical texts and auto-generated corpora capturing the latest developments can further produce pre-trained models to answer some specific questions that commercially available assistants cannot and can serve as intelligent assistants for individual facilities.
\end{abstract}

\section{Introduction}
Modern AI assistants like \chname{ChatGPT}~\cite{chatgpt} and \chname{Claude}~\cite{claudeai} have revolutionized artificial intelligence, showcasing impressive creativity.
However, they rely on questionable internet sources and can generate plausible but incorrect responses when lacking knowledge, well known as hallucinations~\cite{welleck2019neural}, which is a known and limiting feature of LLMs.
Furthermore, the complexity of the numerous commercial large language models makes it difficult to easily fine-tune these models to incorporate the latest advancements.
On the other hand, fine-tuning smaller but powerful LLMs like~\cite{touvron2023llama} or \cite{vicuna2023} for particular domains on more reputable sources like scientific papers addresses these issues.
Scientific papers provide a more trustworthy source of information for creating a more reliable training dataset.
This brings a potentially wide range of applications for the community. For instance, one can fine-tune and automate an LLM to perform department-specific tasks that consist of a specific language.
These intelligent assistants can improve FAIR principles and can help in diverse applications like logbooks which are critical infrastructure components in particle accelerator research and development.
To tackle this, one can consider using off-the-shelf parameter-efficient fine-tuning methods~\cite{hu2021lora,longlora} that adapt smaller models~\cite{vicuna2023,touvron2023llama} into domain or facility-specific assistants that are able to work with complex conceptual questions where commercial chatbots will often fail.
However, the challenging part is the automated preparation of high-quality training data tailored to the domain.

In this work, we present an approach that automates data preparation for fine-tuning publicly available models, enabling training customized AI assistants.
We introduce the first accelerator assistant, trained without a human in the loop in dataset preparation, which can answer questions that commercially available chatbots are not able to answer.
Furthermore, we show an automated pipeline that enables automation for arbitrary domains without the laborious creation of a supervised dataset and automates this process instead.

\begin{figure*}[!t]
    \centering
    \includegraphics[width=\textwidth]{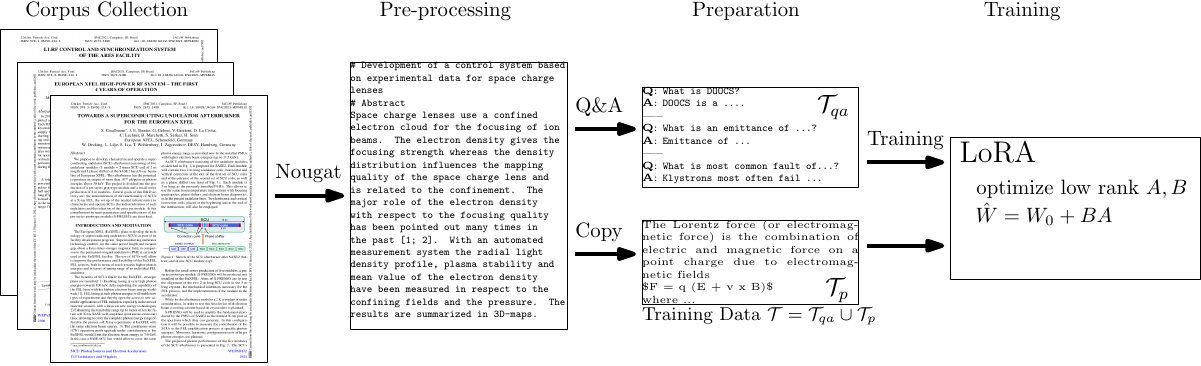}
    \caption{Overview of the training pipeline.
    The shown pipeline compiles complementary text sources (Step 1). These sources are processed with Nougat OCR~\cite{blecher2023nougat} and standardized formats to extract text, equations, and tables (Step 2).
    The structured data is used for text prediction training \Tp{} and to auto-generated question-answer pairs \Tqa{} (Step 3).
    Finally, the prepared dataset fine-tunes an LLM using memory efficient LoRA~\cite{hu2021lora} for robust training (Step 4).}
    \label{fig:pipeline}
\end{figure*}

\section{LLM in Particle Accelerator Technology}
LLMs have the potential to benefit the particle accelerator R\&D community, though applications are not limited to this domain, some examples might be:
\begin{enumerate}
\item LLMs can advance accelerator control and automation by developing systems that respond to condition changes, improving stability and safety, improving the FAIR-ness of logbooks, and helping operators to interpret data, see~\sect{sect:futurework}.
\item LLMs can assist in interpreting facility conditions by harvesting existing (multi-modal) data.
\item A unified LLM could enhance communication and collaboration within and outside facilities by providing natural language interfaces for discussing complex concepts, sharing results, and collaborating on experiments.
\item LLMs can assist in accelerator design and optimization by consolidating knowledge about facilities and making it accessible to personnel, aiding the exploration of configurations and parameters from past or published results via incorporating additional multimodal inputs (images, screenshots, machine parameters), see~\sect{sect:futurework} for more details.
\item LLMs can be used to teach accelerator principles, making this specialized knowledge more accessible to anyone interested.  
\end{enumerate}

\section{Pipeline of Fine-tuning the LLM}
The pipeline can be summarized in the following steps:
Preparing training data begins with collecting corpus sources described in \sect{sect:dataset:sources} consisting of books, conference proceedings, and pre-prints to create a knowledge base in the form of PDF documents.
These sources are processed with optical character recognition (OCR) to extract text, equations, and tables as we explain in \sect{sect:dataset:pre}.
The structured data is directly used for text prediction training (unsupervised dataset \Tp{}). For question answering (supervised dataset \Tqa{}), sections of text are prompted to auto-generate question-answer pairs explained in~\sect{sect:dataset:qg}.
In the final step, in \sect{sect:fine}, the prepared dataset then fine-tunes LLM optimized for long contexts and efficient training, using memory efficient~LoRA~\cite{hu2021lora} to enable robust training on conventional hardware.

The result is an AI assistant tailored to a scientific domain by leveraging accessible training data \T{} spanning current publications and foundational texts. The automated pipeline enables continuous updating to remain updated with the state-of-the-art in the field.
The steps are shown~\fig{fig:pipeline}.

\section{Dataset}
\label{sect:dataset}

\subsection{Corpus Sources}
\label{sect:dataset:sources}
To create a dataset, we need text corpus sources. This work uses three complementary data sources to generate a robust dataset covering foundational knowledge and recent progress. Publicly available books provide broad, established knowledge to build a strong basis, referred to as the \texttt{books} corpus.
Conference proceedings from~\cite{jacow} contain the most up-to-date facility developments, referred to as \texttt{jacow}. arXiv~\cite{arxiv} includes currently peer-reviewed pre-prints in category~\texttt{physics.acc-ph} (accelerator physics) from 2015 till this day and is denoted as \texttt{arxiv}.

The source list, including automated acquisition procedures, will be published with the dataset to enable continuous updates with new publications. This corpus establishes a comprehensive foundation while capturing the latest advancements.

\subsection{Dataset Pre-processing}
\label{sect:dataset:pre}
In the pre-processing step, we focus on digitizing documents into machine-readable formats.

Nougat OCR~\cite{blecher2023nougat} uses a visual transformer to process scientific documents into a computer-readable format. It can transform a scientific paper into computer-readable text, equations, and tables. Nougat converts inputs to MultiMarkdown, which is better suited for training than raw (La)TeX, however since LaTeX is the standard for equation typesetting, we keep the equations in LaTeX.

We transformed all computer-readable outputs from Nougat into a common format with text as Markdown, tables as plain text, and equations in LaTeX. Nougat initially outputs MultiMarkdown, which requires some minor modifications, namely we transformed equations to LaTeX via regular expressions and tables to plain text with Pandoc~\cite{pandoc}.

Only \texttt{books} and \texttt{papers} sources were processed by Nougat. \texttt{arXiv} texts are already in LaTeX, so we used Pandoc~\cite{pandoc} to convert to MultiMarkdown, replaced equation delimiters with LaTeX (\texttt{\$}), and rendered other macros (like image captions) as plain text that are better suited for training.

\subsection{Dataset Preparation}
\label{sect:dataset:qg}
The unsupervised dataset \Tp{} for the text prediction does not require any further steps after pre-processing and we pass the input data as we describe in the pre-processing step, see \sect{sect:dataset:pre}.

To generate question-answer supervised pairs we prompted~\chname{vicuna-7b-16k-v1.5} with a following prompt: \texttt{Generate ten questions for a paper:"\$TEXT"} where \texttt{\$TEXT} is the actual paper to generate ten questions per either a section (or subsection) form \texttt{book} corpus or paper from \texttt{jacow} corpus.
If the generated answer did not follow a pre-specified format, it was discarded.

The entire dataset consists of 633759 samples, from which 24949 are unsupervised \texttt{arXiv} samples (papers), 1689 samples are from \texttt{books} corpus (sections or subsections, depending on length), and 338207 samples (papers) are from \texttt{papers} corpus. Generated supervised question-answer pairs consist of 13705 samples from \texttt{book} (single question and answer pair, if processed properly) corpus and 255209 \texttt{papers} corpus (single question and answer pair, if processed properly).

\section{Fine-tuning}
\label{sect:fine}
We leverage the pre-trained transformer language model \chname{vicuna-7b-16k-v1.5}~\cite{vicuna2023}. It provides a robust foundational model while being reasonably small (7 billion parameters) and is trainable on conventional GPUs. Furthermore, it can handle up to 16k input tokens, essential for often lengthy scientific texts.
Compared to larger models like \chname{llama-70b}~\cite{touvron2023llama}, it has fewer parameters and thus it is easier to be continuously updated and fine-tuned on conventional hardware.
Fine-tuning uses LoRA~\cite{hu2021lora} to make large language model training accessible on standard hardware. This combination provides a capable foundation while remaining efficient and accessible.
Considering the input lengths, one may consider saving computational time by using LongLoRA~\cite{longlora} in the future.
We set the training LoRA rank to 64 since datasets contain some intricate texts like equations and tables, $\alpha$ is set to 128. We optimized the query, key, value, and projection weights.
Per-device train batch size set to 2 and gradient accumulation 16 turned out to be a good trade-off between training time and model efficiency.
The training was performed with community-suggested 4 epochs and with a learning rate $5\times 10^{-5}$. The training was done on NVIDIA A100.

\section{Evaluation}
We evaluated our model with state-of-the-art open-source \chname{falcon-180b} AI~\cite{falcon} and \chname{ChatGPT}~\cite{chatgpt}. The shown results are on the model which was trained on papers that contain mention of DESY. We also publish a model that is trained on the entire \T{}. When the output didn't contain a definitive answer, we omitted the text, indicated by ellipsis.

\vspace{-1.2em}
\paragraph{Question 1 about DESY ARES} : \chat{Does ARES use normal or superconducting cavities?}

\vspace{-0.5em}
\chname{PACuna}: \chat{ARES uses normal cavities for its accelerator}, \correct{Correct}

\vspace{-0.5em}
\chname{falcon-180b}: \chat{I do not have information about a specific ares system. However, ares is a general acronym that can refer to various systems...}. \wrong{Wrong, the answer is missing.} \\

\vspace{-1.2em}
\chname{ChatGPT}: \chat{As of my last knowledge update in September 2021, ARES (Accelerator for Research in Subatomic Physics) had plans to use superconducting radiofrequency (SRF) cavities ...}, \wrong{Wrong, they are not superconducting.}

\vspace{-1.2em}
\paragraph{Question 2 about DESY European XFEL}: \chat{How many superconducting radio frequency (SRF) cavities does European XFEL have?}

\vspace{-0.5em}
\chname{PACuna}: \chat{European XFEL has a total of 800 superconducting radio frequency (SRF) cavities.}, \correct{Approximately correct, it is 808.}

\vspace{-0.5em}
\chname{falcon-180b}: \chat{The European XFEL has a total of 16 superconducting radio-frequency (SRF) cavity stations that contain approximately 170 individual SRF cavities ...}, \wrong{Wrong.}

\vspace{-0.5em}
\chname{ChatGPT}: \chat{As of my last knowledge update in September 2021, the European XFEL (X-ray Free-Electron Laser) facility had a total of 100 superconducting radiofrequency (SRF) cavities ...}, \wrong{Wrong, European XFEL has 100 modules.}

\vspace{-1.5em}
\paragraph{Question 3 about beam emittance at REAGE}: \chat{What is the significance of measuring the beam emittance in the REGAE linear accelerator?}

\vspace{-0.5em}
\chname{PACuna}:\chat{Measuring the beam emittance in the REGAE linear accelerator is significant because it enables the calculation of the beam's emittance, which is a measure of the beam's transverse size and intensity. This information is crucial for optimizing the performance of the REGAE linear accelerator and the other devices it serves.}, \correct{Partially correct, word intensity is not correct.}

\vspace{-0.5em}
\chname{falcon-180b}: \chat{Measuring the beam emittance in the REGAE linear accelerator is significant because it allows for the optimization of the accelerator's performance. Emittance is a measure of how tightly packed the particles in the beam are, and a smaller emittance means a more focused and uniform beam...}, \correct{Correct.}

\vspace{-0.5em}
\chname{ChatGPT}: \chat{Measuring the beam emittance in a linear accelerator like REGAE (Relativistic Electron Gun for Atomic Exploration) is of significant importance because emittance is a fundamental parameter that characterizes the quality of the charged particle beam. The emittance measurement provides crucial information about the beam's properties, and it has several significant implications for accelerator operation and scientific experiments...}, \correct{Correct.}

Experiments show the power of fine-tuning facility-specific questions, which is the main strength of our model. The main strength is the model's ability to answer facility-specific questions (Questions 1, 2), which is understandable considering the generality of the larger models.
However, the other models provided a very good explanation for the third question, where \chname{PACuna} provided misleading information.

\section{Future Work}
\label{sect:futurework}
In the future, we plan to expand our data sources to incorporate more relevant and text-rich sources like~\cite{Moskovic:2021zjs}, while simultaneously enhancing our data preparation techniques for the sources already used.
Notably, some sources were regrettably excluded due to the absence of source files in the case of \texttt{arXiv} corpus. This can be mitigated by applying the same processing methodology successfully used for the~\cite{jacow} papers.  
Additionally, we aim to generate more intricate and nuanced questions through specialized techniques like~\cite{ushio2023toolkit}, refining existing prompts, or instruction mining~\cite{cao2023instruction}.
Currently, our generated questions are robust but generic since we always generate a fixed number of question-answer pairs per input, which may lead to ignoring some knowledge.
We also plan a more detailed evaluation in future work with a human expert in the loop in the assessment of the generated data including hallucinations.
What we found particularly crucial is preventing hallucinations, therefore in the future we would like to provide countermeasures against it either by scoring the answer or giving a source~\cite{tian2023fine}.

Recently, work combining text, LLMs, and multi-modal data like images~\cite{wang2023visionllm} or even point sets~\cite{xu2023pointllm} has emerged.
This is particularly interesting for particle accelerators which generate large amounts of multi-modal data. AI assistants then can for instance help with searching parameters (from point sets) or interpret the descriptions and images from logbooks.

\section{Conclusion}
\label{sect:conclusion}
In conclusion, this work demonstrates an approach for automating the fine-tuning of language models for specialized domains, using particle accelerator technology as an example application. We introduced \chname{PACuna}, an LLM tailored for the particle accelerators through automated fine-tuning on a large corpus of domain-specific texts.
The corpus was compiled from publicly available sources like books, conference proceedings, and pre-prints, providing both foundational knowledge and cutting-edge developments. Equations and tables were extracted and converted to machine-readable formats with~\cite{blecher2023nougat} and question-answer pairs were automatically generated to create training data for fine-tuning.
The model shows reasonable performance in addressing highly facility-specific problems in particle accelerators, as we show in the evaluation section, and is often on par with theoretical knowledge available with the pre-trained assistants.

The shown methodology provides a generalizable framework for creating domain-specific LLMs, overcoming some limitations of the general-purpose chatbots.
This contributes a proof-of-concept for the feasibility of automated fine-tuning to align LLMs with rapidly advancing scientific fields. By concentrating knowledge and reasoning skills, PACuna and models derived through similar approaches can enhance accelerator R\&D, control systems, collaboration, and education.

\section*{Acknowledgement}
We acknowledge DESY (Hamburg, Germany), a member of the Helmholtz Association HGF, for its support in providing resources and infrastructure. Furthermore, we would like to thank all colleagues of the MCS and MSK groups and management for their contributions to this work and help in preparing this paper.

\bibliographystyle{plainnat}
\bibliography{references}

\clearpage
\section*{Checklist}


\begin{enumerate}

\item For all authors...
\begin{enumerate}
    \item Do the main claims made in the abstract and introduction accurately reflect the paper's contributions and scope? \answerYes{}
    \item Did you describe the limitations of your work? \answerYes{}
    \item Did you discuss any potential negative societal impacts of your work? \answerNo{}
    \item Have you read the ethics review guidelines and ensured that your paper conforms to them? \answerYes{}
\end{enumerate}

\item If you are including theoretical results...
\begin{enumerate}
    \item Did you state the full set of assumptions of all theoretical results? \answerNA{}
	\item Did you include complete proofs of all theoretical results? \answerNA{}
\end{enumerate}

\item If you ran experiments...
\begin{enumerate}
  \item Did you include the code, data, and instructions needed to reproduce the main experimental results (either in the supplemental material or as a URL)? \answerYes{}
  \item Did you specify all the training details (e.g., data splits, hyperparameters, how they were chosen)? \answerYes{}
  \item Did you report error bars (e.g., with respect to the random seed after running experiments multiple times)? \answerNo{At present, we have not encountered a benchmark dataset in the published work that aligns closely with the aims of our paper.}
  \item Did you include the total amount of compute and the type of resources used (e.g., type of GPUs, internal cluster, or cloud provider)? \answerYes{}
\end{enumerate}

\item If you are using existing assets (e.g., code, data, models) or curating/releasing new assets...
\begin{enumerate}
  \item If your work uses existing assets, did you cite the creators? \answerYes{}
  \item Did you mention the license of the assets? \answerNo{}
  \item Did you include any new assets either in the supplemental material or as a URL? \answerYes{}
  \item Did you discuss whether and how consent was obtained from people whose data you're using/curating? \answerNo{Most sources are publicly available on-line and we compile a list of sources on the paper's website.}
  \item Did you discuss whether the data you are using/curating contains personally identifiable information or offensive content? \answerNo{Retaining names, affiliations, and contact details in training data upholds norms of scholarly attribution and enables potential collaboration between researchers.}
\end{enumerate}

\item If you used crowdsourcing or conducted research with human subjects...
\begin{enumerate}
  \item Did you include the full text of instructions given to participants and screenshots, if applicable? \answerNA{}
  \item Did you describe any potential participant risks, with links to Institutional Review Board (IRB) approvals, if applicable? \answerNA{}
  \item Did you include the estimated hourly wage paid to participants and the total amount spent on participant compensation? \answerNA{}
\end{enumerate}

\end{enumerate}

\end{document}